%% file: iclr2025_conference.tex
\documentclass{article} 
\usepackage{iclr2025_conference,times}
\usepackage{graphicx}
\input{math_commands.tex}

\usepackage{array}
\usepackage[colorinlistoftodos]{todonotes}
\usepackage{hyperref}
\usepackage{url}
\usepackage[utf8]{inputenc}
\usepackage{algorithm}
\usepackage{booktabs} 
\usepackage{algorithmic} 

\title{\vspace{-0.65cm}Text to 3D Object Generation for Scalable Room Assembly\vspace{-0.2cm}}

\author{Sonia Laguna \thanks{Research conducted during an internship at Google. Correspondence to \texttt{slaguna@inf.ethz.ch}.} \\
ETH Zurich\\
\And
Alberto Garcia-Garcia \\
Google \\
\And
Marie-Julie Rakotosaona \\
Google \\
\And
Stylianos Moschoglou \\
Google \\
\And
Leonhard Helminger \\
Google \\
\And
Sergio Orts-Escolano \\
Google \\
}

\iclrfinalcopy 
\begin{document}

\maketitle
\vspace{-0.5cm}
\begin{abstract}
Modern machine learning models for scene understanding, such as depth estimation and object tracking, rely on large, high-quality datasets that mimic real-world deployment scenarios. To address data scarcity, we propose an end-to-end system for synthetic data generation for scalable, high-quality, and customizable 3D indoor scenes. By integrating and adapting text-to-image and multi-view diffusion models with Neural Radiance Field-based meshing, this system generates high-fidelity 3D object assets from text prompts and incorporates them into pre-defined floor plans using a rendering tool. By introducing novel loss functions and training strategies into existing methods, the system supports on-demand scene generation, aiming to alleviate the scarcity of current available data, generally manually crafted by artists. 
This system advances the role of synthetic data in addressing machine learning training limitations, enabling more robust and generalizable models for real-world applications.
\end{abstract}
\vspace*{-0.5cm}
\section{Introduction and Background}

Synthetic data has become a cornerstone in the training and development of machine learning (ML) models, particularly for perception tasks in computer vision. Applications spanning autonomous navigation, indoor scene understanding, and robotic interaction rely on vast amounts of high-quality, annotated training data that reflects real-world complexity~\citep{nikolenko2021synthetic}. However, acquiring such data at scale poses significant challenges, including privacy and fairness concerns, high annotation costs, and domain-specific constraints. Synthetic data generation offers an efficient and scalable solution, enabling the creation of diverse, controlled datasets that enhance model generalization to real-world environments. In the context of scene understanding, synthetic 3D indoor environments play a crucial role. Realistic room layouts populated with diverse and customizable assets provide valuable training grounds for perception tasks such as depth estimation, object segmentation, and scene reconstruction~\citep{roberts2021hypersim, zheng2020structured3d}. By simulating variations in object type, color, placement, and material properties, synthetic data can bridge domain gaps that arise from the limited variability of real-world datasets. This ability to generalize across domains is pivotal for deploying perception models in unpredictable real-world scenarios.

Existing approaches to synthetic data generation and representation for 3D scenes span a wide range of methodologies, including procedural content creation~\citep{raistrick2024infinigen}, diffusion-based multi-view approaches~\citep{gao2024cat3d}, Neural Radiance Fields (NeRFs)~\citep{mildenhall2021nerf}, 3D Gaussian Splatting~\citep{kerbl20233d}, and 3D supervised latent generative models~\citep{honglrm}.
Meshes in 3D modeling are essential for bridging the mentioned abstract 3D representations with practical applications. By defining geometry through vertices, edges, and faces, they enable realistic rendering, efficient texture mapping, and seamless integration into graphics pipelines. Their lightweight structure ensures real-time performance, making them crucial for AR/VR, gaming, virtual prototyping, and scientific visualization. As demand for photorealistic 3D content grows, high-quality meshes remain vital for immersive and interactive experiences. This deepens the need for synthetic data in the form of 3D meshes. However, generating complete meshes of 3D scenes remains a challenge \citep{hollein2023text2room}.  We include an extensive description of prior related work in Appendix~\ref{app:rel_work}. In this work, by focusing on single asset generations in combination with existing synthetic scenes, we introduce a system to leverage prior synthetic data and scale up 3D indoor scene generations in a mesh format from a text prompt.

\begin{figure}[h!]
\includegraphics[width=\textwidth]{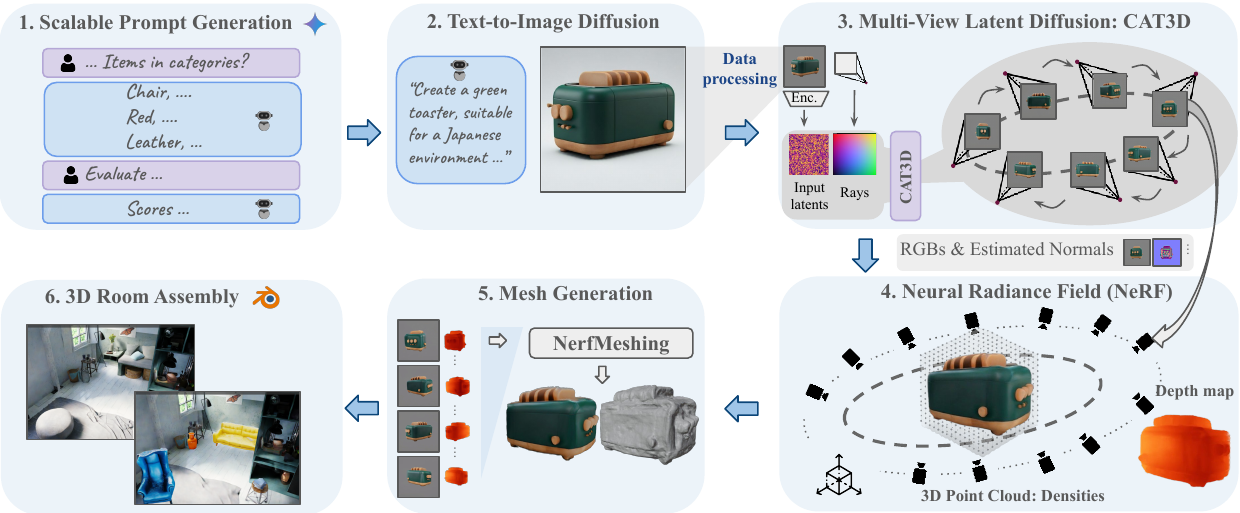}
\vspace*{-0.5cm}
\caption{Overview of the proposed system from scalable prompt generation (\textbf{1})-Section~\ref{sec:methods1}, through text-to-image diffusion (2)-Section~\ref{sec:methods2}, multi-view latent diffusion (3)-Section~\ref{sec:methods3}, and NeRF (4)-Section~\ref{sec:methods4}, onto the resulting meshes (\textbf{5})-Section~\ref{sec:methods5} integrated in existing rooms (\textbf{6})-Section~\ref{sec:methods6}.}
\label{fig:schema}
\end{figure}

We introduce a hybrid system for scalable and customizable 3D room generation, combining pre-existing floor plans with prompt-engineered 3D asset creation. This approach enhances diversity, fidelity, and scalability while reducing reliance on manually designed datasets, supporting robust downstream ML models for perception tasks. By addressing common challenges, such as background speckles, global 3D consistency, and geometry fidelity, we incorporate refined prompt engineering, instance segmentation, object-based fine-tuning, and sparsity and normal regularization. Our key contributions are: (\emph{i})~We propose an automated prompt generation system that scales synthetic asset creation while ensuring high diversity and consistency for indoor scenes. (\emph{ii})~We extend and integrate state-of-the-art techniques in multi-view generation, NeRF representation, and meshing. Our improvements include fine-tuning diffusion models, applying sparsity loss functions, enhancing background segmentation, and leveraging advanced monocular depth estimation for precise surface normal supervision. (\emph{iii})~We introduce an end-to-end system that integrates generated assets into artist-defined room layouts. We evaluate the synthesized environments showing their potential to bridge the gap between synthetic and real-world data, setting a new benchmark for synthetic scene generation in ML perception tasks. Figure~\ref{fig:schema} illustrates an overview of the proposed system.

\vspace*{-0.3cm}
\section{Methods}
\label{sec:methods}

The system integrates existing and adapted methods into a pipeline with subcomponents, as seen in Figure~\ref{fig:schema}. This modular architecture grants flexibility to adapt to future advancements in ML research. The following section provides a detailed description of the components.

\subsection{Prompt Engineering and Scalability}
\label{sec:methods1}
Prompt engineering plays a critical role in generating high-quality synthetic 3D scenes. Automated prompt generation expands object diversity in the synthetic data by incorporating domain-specific details. This involves designing precise textual inputs to ensure outputs meet desired specifications, including object type, material, color, and style. The nuances of prompt engineering significantly influence the quality, coherence, and scalability of generated assets.

\paragraph{Contextual Precision}  
We conducted a study including specific contextual phrases to refine the text prompts. We found that adding in ``in an empty white background'' and ``in the middle'' improved spatial clarity, preventing issues like object cropping or awkward placements. These ensured that generated assets align with the pipeline's requirements, as shown in Figure~\ref{fig:results_cont_preci} in Appendix~\ref{app:prompts}.

\paragraph{Scalable Prompt Generations}

We introduce a procedure for automating the generation of diverse, high-quality text prompts for synthetic room assets: (i) \textit{Gemini} is used to generate category lists for objects, materials, colors, and high-level themes, each with 10 to 20 realistic entries, as described in Steps 1 and 2 of Algorithm~\ref{alg:dataset_creation}. For instance, objects might include chairs, tables, and vases, while materials range from wood and metal to fabric and glass. Descriptive colors such as blue or  metallic gray further diversify the dataset. (ii) \textit{Gemini} creates natural language templates (Step 3) to integrate these categories in various sentence structures, avoiding redundancy. Examples include: ``Design a [Color] [Material] [Object] inspired by [High-level Theme] aesthetics,'' or ``Create a [Object] that is both functional and aesthetically pleasing, using [Color] [Material] within a [High-level Theme] setting.'' (iii) Next, we combine the generated categories and templates to create diverse prompts by permuting the variables (Step 4). (iv) \textit{Gemini} evaluates each prompt based on coherence, specificity, and creativity (Step 5), to ensure the prompts are realistic, plausible, and visually inspiring. (v) Finally, we rank the prompts based on the final score (Step 6) and save them as a \textit{csv} file to feed them to the next step. This corresponds to Part 1 as shown in Figure~\ref{fig:schema}. A comprehensive explanation of the procedure is included via pseudocode in Algorithm~\ref{alg:dataset_creation}, and Table~\ref{tab:prompt_evaluation} shows examples of the different ranges of scores of the resulting prompts with each resulting generated mesh, both in Appendix~\ref{app:prompts}.   

\subsection{Text-to-Image Generation}
\label{sec:methods2}
This section builds upon previous prompt engineering by utilizing state-of-the-art text-to-image diffusion models,
to generate high-quality 2D images for 3D reconstruction pipelines. Text prompts allow for scalable, adaptable, and diverse synthetic data generation, reducing reliance on real-world captures while offering fine-grained control over content (see part 2 in Figure~\ref{fig:schema}). Furthermore, we integrate a segmentation step using DeepLab~\citep{chen2017deeplab} to isolate objects and ensure smooth, homogeneous backgrounds, as well as adding a background with a padding of 20\% around the object, key for isolated object generation. This design choice in particular addresses common issues in text-to-image models when the goal is to generate single objects. Generally, extraneous elements or cluttered backgrounds are present and can interfere. Segmentation improves the quality and consistency of inputs for 3D isolated object reconstruction. This refinement step is visualized as data processing between Part 2 and Part 3 in Figure~\ref{fig:schema}.

\subsection{Multi-View Diffusion}
\label{sec:methods3}
We build upon state-of-the-art multi-view latent diffusion model (LDM) approaches to generate globally consistent views. In particular, we focus on CAT3D~\citep{gao2024cat3d}, which has demonstrated superior performance in the literature. Prior models offering multi-view outputs like MVDream~\citep{shi2023mvdream}, ImageDream~\citep{wang2023imagedream}, SyncDreamer~\citep{liu2023syncdreamer}, SPAD~\citep{cusini2022historical}, SV3D~\citep{voleti2024sv3d}, and Zero123++~\citep{shi2023zero123++} could be used instead to generate the image-to-3D assets. The CAT3D framework, in particular, extends traditional diffusion models by introducing a multi-view diffusion process that generates consistent 2D projections from single-image inputs and camera poses. Additionally, an autoregressive sampling of the views ensures geometric alignment across views, which is critical for accurate 3D reconstruction. 

CAT3D addresses the challenge of generating coherent views from sparse inputs through the use of a multi-view latent diffusion model. The approach is structured around three core processes. First, latent representation encoding is performed by employing a variational auto-encoder (VAE) to transform input images into a latent space, capturing both spatial and semantic features necessary for reasoning about unseen views. Second, pose-conditioned diffusion ensures the model is conditioned on camera poses corresponding to observed and target viewpoints, achieved by concatenating a relative raymap representation of the camera's position and orientation with the latent embeddings. Third, progressive view synthesis is executed, where the model generates multiple target views by iteratively refining noisy latents into high-fidelity projections, thereby ensuring spatial consistency across synthesized outputs, as illustrated in Part 3 in Figure~\ref{fig:schema}.

Consistency across synthesized views is paramount for downstream 3D reconstruction tasks. To this end, CAT3D incorporates innovations such as 3D self-attention layers and binary masks to indicate observed inputs during training, addressing the inherent challenges of sparse input scenarios. Strategic anchor views are selected using an autoregressive sampling strategy, enabling efficient reasoning from limited data. To scale to dense viewpoints, the model clusters proximal camera positions and leverages anchor views to maintain coherence across hundreds of target viewpoints. Robustness to sparse inputs is further enhanced by advanced loss functions, including LPIPS perceptual loss, which preserves semantic integrity even in conditions of high uncertainty. CAT3D generates all kinds of 3D scenarios and multi-object scenes. However, full scenes pose a big challenge with the objective of generating 3D object mesh outputs. Therefore, in the system, we fine-tune CAT3D with internal high-quality single object datasets, specified in Section~\ref{sec:exp_setup}, to avoid background speckles or inconsistencies when shifting the focus from scene-based data.

\subsection{NeRF to Create 3D Representations}
\label{sec:methods4}
\begin{figure}[h!]
\includegraphics[width=\textwidth]{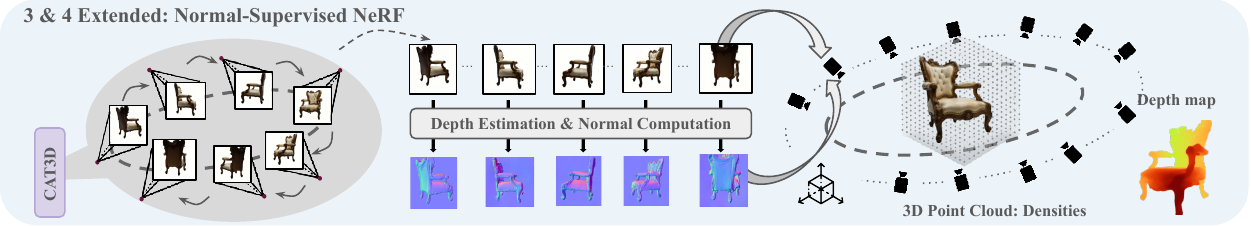}
\vspace*{-0.2cm}
\caption{Normal estimation and its use in the NeRF, expanding on Steps 3 and 4 from Figure~\ref{fig:schema}.}
\label{fig:schema_normal}
\end{figure}

NeRFs transform multi-view 2D projections into detailed 3D scene representations. They are neural network-based models that generate high-quality 3D scenes from 2D images by learning to represent volumetric density and view-dependent color at any 3D point, as illustrated in Part 4 in Figure~\ref{fig:schema}. Our approach inherits ZipNeRF~\citep{barron2023zip} with the modifications done for CAT3D and builds on it with novel enhancements, which allow for more robust and photorealistic 3D reconstruction focused on single objects, even from sparse data. These are: \textit{Density Regularization} for better geometric fidelity to a single object, by including a logarithmic sparsity loss on the output densities ($\sigma$), ${L_d} = \text{mean}(\log(\sigma))$; \textit{Input Segmentation} of all the input views using DeepLab for background consistency; \textit{Normal Smoothness and Orientation Losses}, as introduced originally in~\citet{verbin2022ref}; and \textit{Estimated Normal Supervision}, by leveraging MariGold (MG)~\citep{gupta2022marigold}, a state-of-the-art monocular depth estimation model. This last step uses the estimated depth to compute its gradients and use them as view-corresponding normals to regularize the geometry of the resulting asset. In particular, we compute the cosine loss on the estimated normal from the NeRF density, $\hat{\mathbf{n}}$, and the normal predicted from the MG approximation, $\hat{\mathbf{n}}^{\text{MG}}$.
Note that by leveraging existing models we alleviate the need to train with geometry-informed 3D data. Figure~\ref{fig:schema_normal} shows this step in a more comprehensive manner with the visualization of the predicted normals. 
    
\subsection{NeRFMeshing: Distilling NeRF into Meshes}
\label{sec:methods5}
NeRFMeshing~\citep{rakotosaona2024nerfmeshing} is a method designed to extract geometrically accurate 3D meshes from NeRFs, enabling their integration into standard computer graphics and simulation pipelines. By introducing a novel Signed Surface Approximation Network (SSAN), it distills the volumetric representation of NeRFs into a compact mesh format with precise geometry and view-dependent color properties. This process facilitates real-time rendering and supports physics-based simulations, overcoming traditional NeRF limitations in geometry accuracy and compatibility with graphics workflows. We use NeRFMeshing with additional (i) SSIM and (ii) LPIPS supervision for improved rendering after the mesh extraction.

\subsection{Application: Placing Generated Objects in Rooms}
\label{sec:methods6}
We demonstrate the scalability of our system by incorporating generated 3D assets into complex room layouts using Blender, open source rendering software. The process includes: \textit{Asset Scaling and Placement} by importing assets for intuitive scaling and alignment and \textit{Scene Composition}, by positioning objects to achieve realistic spatial arrangements. This supports creating realistic, editable scenes for applications like simulation, architectural visualization, and interactive media.

\section{Experimental Setup}

\label{sec:exp_setup}
\textbf{Dataset:} The data used to finetune CAT3D consists of internal high-quality datasets containing more than 300K isolated objects. These datasets highlight synthetic data’s capacity to generalize across domains. Furthermore, we apply the system to place objects in more than 64 rooms (crafted by 3D artists from Evermotion) in Blender.


\textbf{Resources:} The system uses 16 A100 GPUs to sample the multi-view diffusion model and to train and evaluate the NeRF and NerfMeshing architectures in up to 15 minutes. Note that prior work on mesh asset generation, i.e. TextMesh~\citep{tsalicoglou2024textmesh}, takes 2 hours to generate a single asset due to SDS optimization. Our system allows for flexibility of generations based on a prompt, as most steps have not been trained on specific 3D datasets, unlike prior models trained end-to-end with 3D-geometry-informed supervision that can struggle to generate out-of-domain when the input prompts are diverse. Our model is implemented in Jax v0.5 and room generation in Blender 4.3.

\section{Results}

Our results highlight the effectiveness of our approach in 3D asset generation and scene composition, demonstrating geometric consistency, texture quality, and overall realism. Comprehensive visual evaluations establish our method as a new benchmark in the field. 

\subsection{Effect of Object Segmentation, Fine-tuning and Sparsity}

To evaluate the effect of object segmentation, fine-tuning of CAT3D, and density regularization, we compare results generated by our system without these components against those generated with their inclusion. Without them, the generated assets lack precise boundaries and exhibit blending artifacts with the background, we see a significantly enhancement of the fidelity of textures and geometric details. Visual comparisons that illustrate the marked improvement in multi-view consistency and realism are displayed in Figure~\ref{fig:examples_ablations}a.

 \subsection{NeRF and NerfMeshing Geometry Adaptations}
 
\begin{figure}[h!]
\centering
\includegraphics[width=\linewidth]{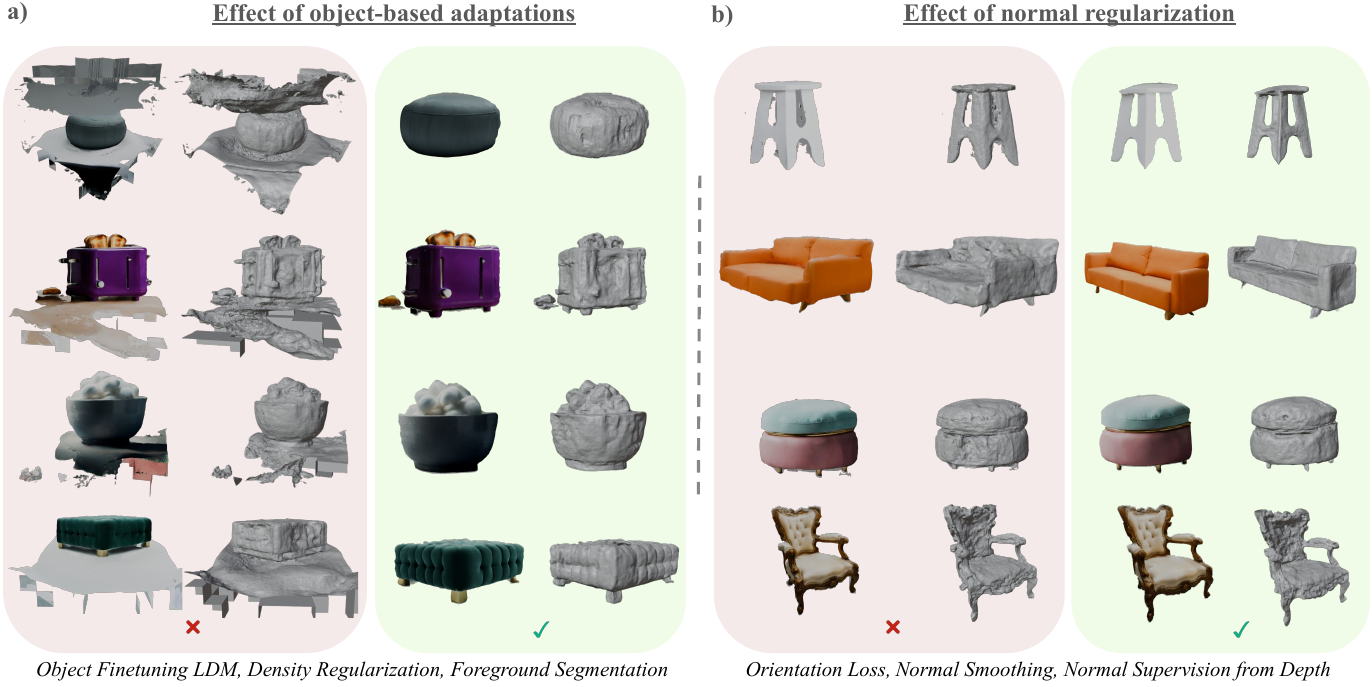}
\caption{Qualitative evaluation of the contributions of our system to the different subcomponents. 
(a) Impact of object-based adaptations including segmentation, fine-tuning of the LDM, and density regularization. 
(b) Effect of NeRF and NeRFMeshing adaptations focusing on normal regularization. In particular, adding normal smoothness, orientation loss, and normal supervision.}

\label{fig:examples_ablations}
\end{figure}

 To evaluate the impact of our introduced modifications to NeRF, particularly the normal smoothness, orientation loss and normal supervision, we visualize assets generated from the same prompt with and without this components, as shown in Figure~\ref{fig:examples_ablations}b. The introduced changes reduce artifacts and improve the quality of reconstructions, in particular visible with improved smoothness at the edges of the 3D meshes and in complex or sharp regions.

\subsection{Qualitative Results}
\begin{figure}[h!]
\centering
\includegraphics[width=0.4\linewidth]{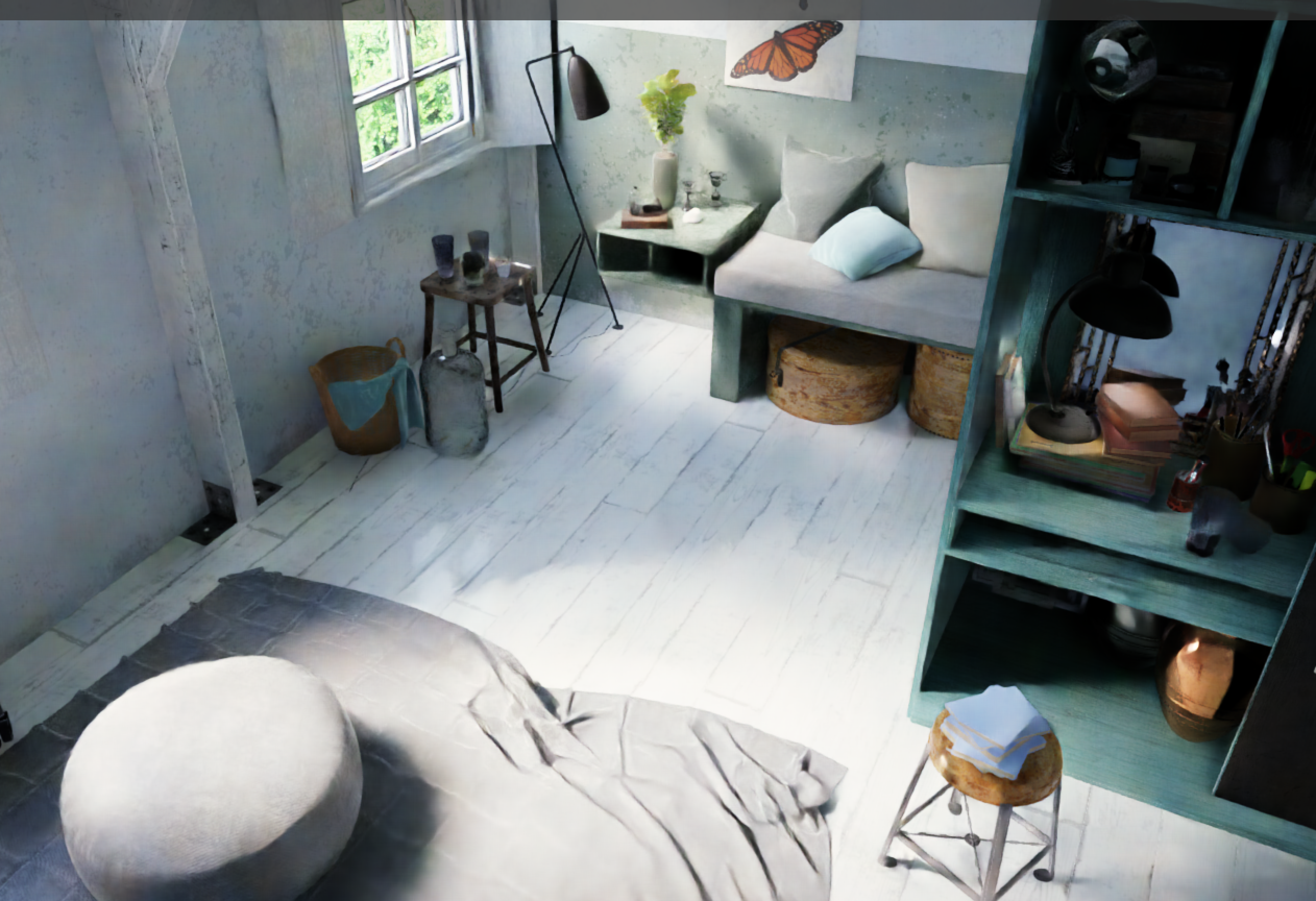}
\includegraphics[width=0.4\linewidth]{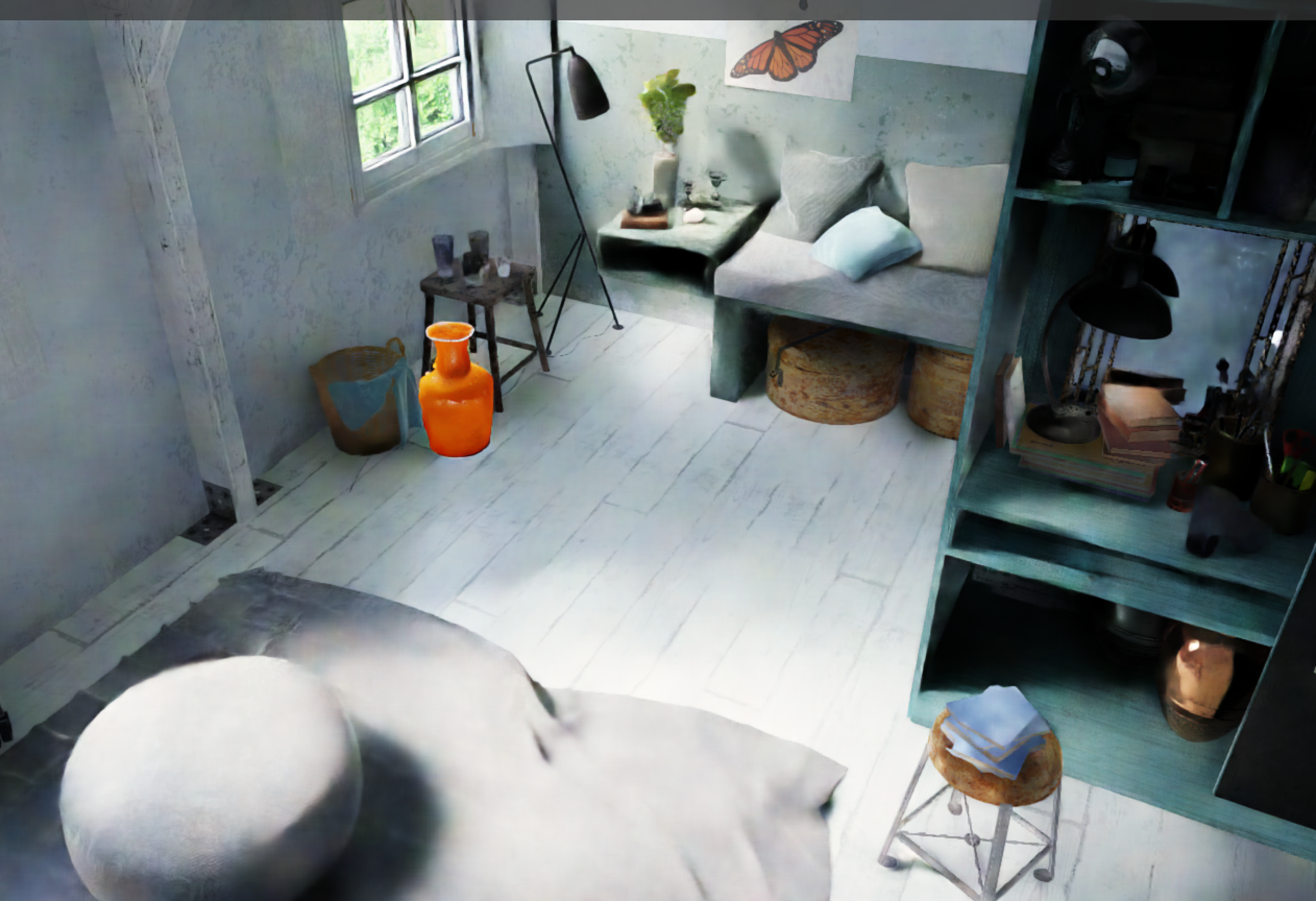}\\
\includegraphics[width=0.4\linewidth]{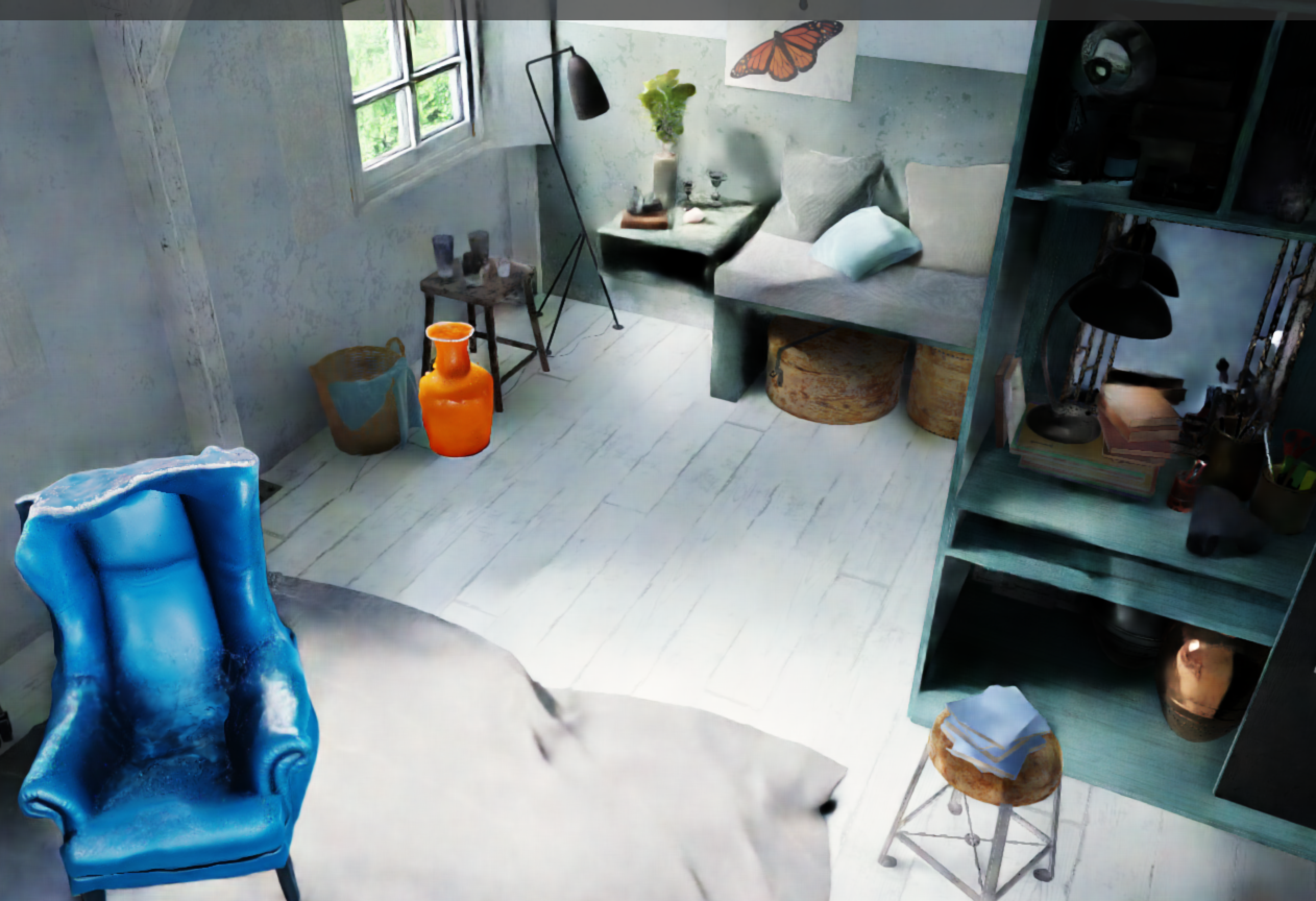}
\includegraphics[width=0.4\linewidth]{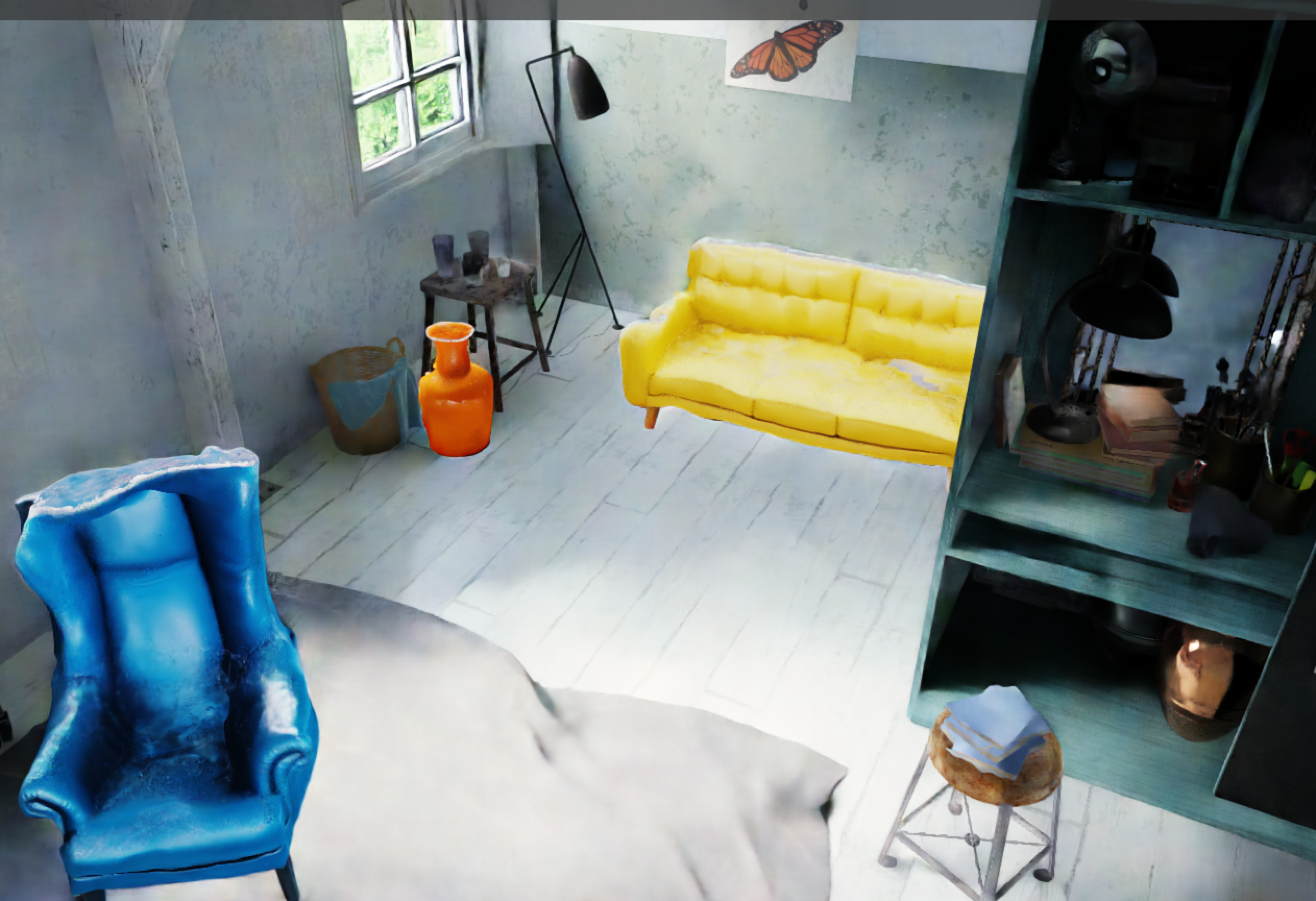}\\
\includegraphics[width=0.4\linewidth]{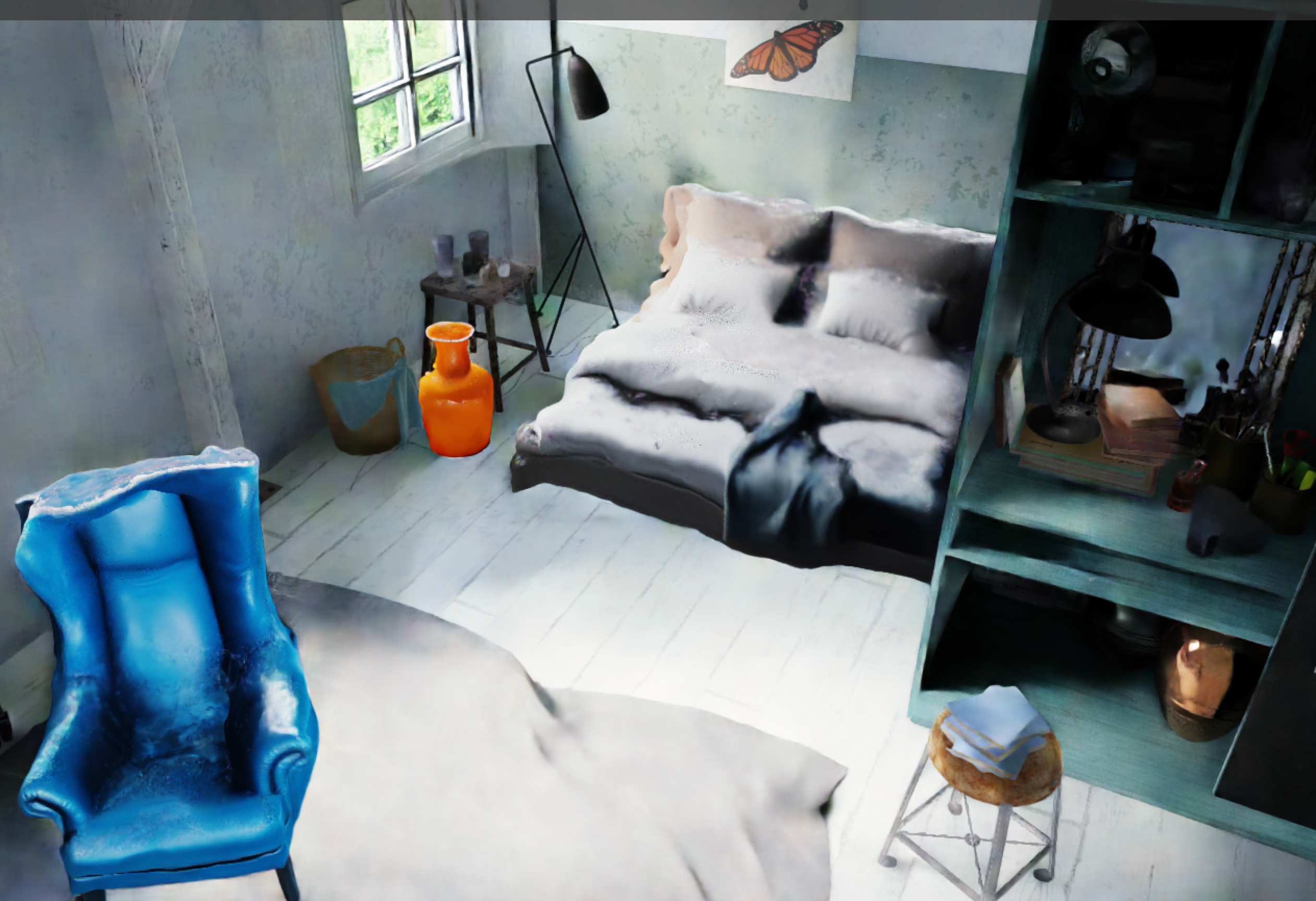}
\includegraphics[width=0.4\linewidth]{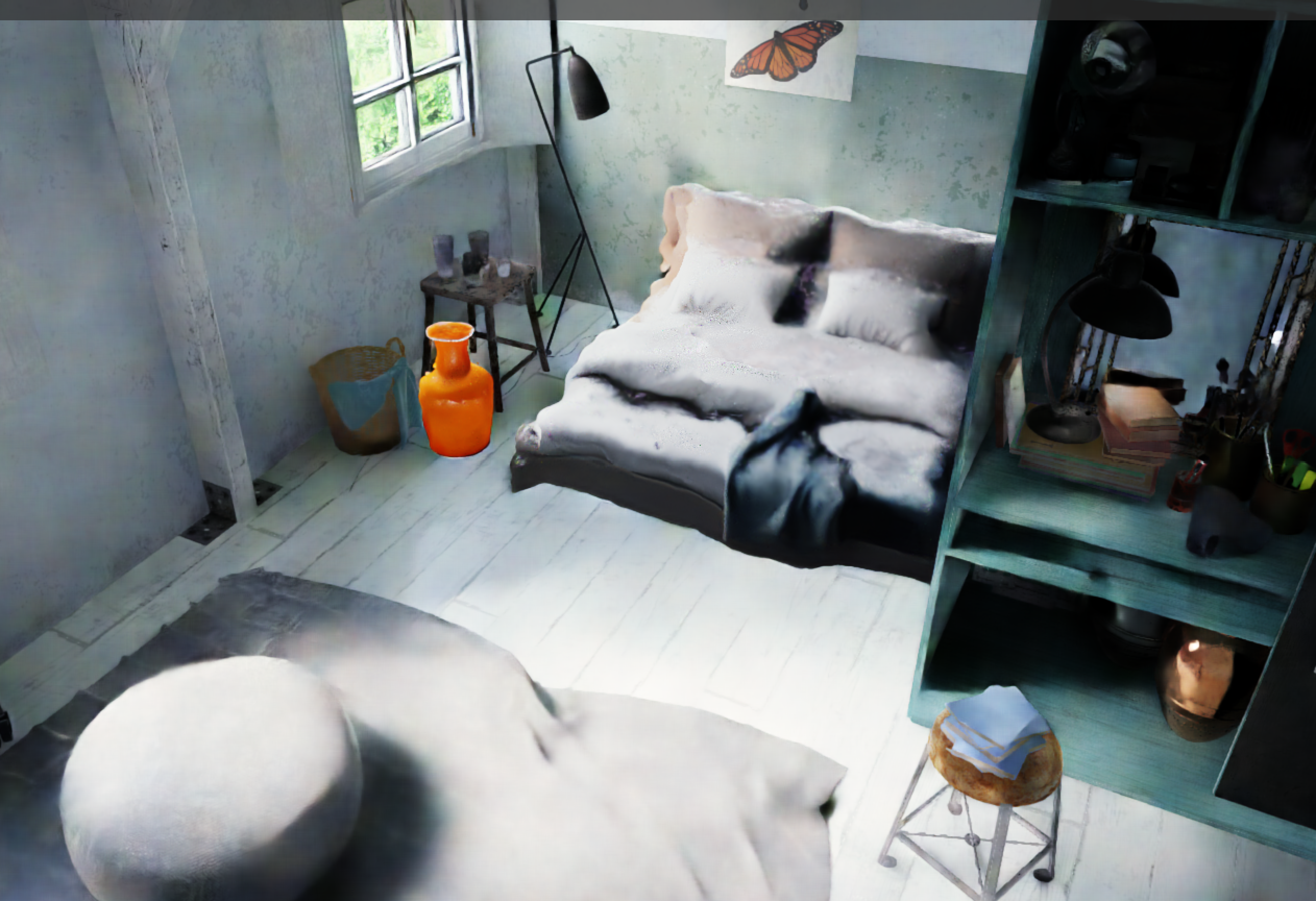}
\vspace*{-0.1cm}
\caption{Synthetic 3D room variations generated by the system. From the original room (top left), annotated objects are replaced by generated semantic equivalents to produce room permutations.}

\label{fig:examples}
\end{figure}

To showcase the results of the complete end-to-end system, Figure \ref{fig:examples} illustrates the results of Step 6 in Figure~\ref{fig:schema}, the 3D Room Assembly. It displays an example starting from an original room (top left), where sequentially, annotated objects are replaced by semantic equivalents generated with the proposed system, to produce permutations of the same room. Further results for individually generated objects and failure modes have been included in Figure~\ref{fig:examples_assets} in Appendix~\ref{app:results}.

\vspace*{-0.1 cm}
\section{Discussion and Conclusion}
\vspace*{-0.15cm}

We introduce an automated system that successfully generates high-fidelity 3D assets from text descriptions, leveraging state-of-the-art diffusion models, NeRF-based meshing, and advanced prompt engineering. The approach shows object consistency and realism while enabling scalable synthetic data generation for downstream ML and perception applications. Key contributions include fine-tuned multi-view diffusion, object segmentation, and normal estimation supervision, all of which significantly improve geometric accuracy and visual fidelity, as shown through visualizations. We show how these assets can be placed and ensure seamless integration into existing environments. Its relevance lies in addressing the key challenge of generating 3D models, which stems from the limited availability of 3D training data. This limitation results in significantly smaller datasets compared to those used in other domains like 2D image synthesis. Despite these advancements, challenges remain in handling complex multi-object scenes and automated object placement. Future work could focus on optimizing spatial arrangements. Our system controls synthetic data creation, bypassing privacy and copyright concerns, and sets a new benchmark in synthetic scene assembly and generation, bridging the gap between real and artificial data, with applications in virtual reality, gaming, and AI-driven simulation.

\section*{Acknowledgements}
We thank the GLDM team at Google for their valuable contributions. 

\bibliography{iclr2025_conference}
\bibliographystyle{iclr2025_conference}
\newpage
\appendix
\section{Related Work}
\label{app:rel_work}

\paragraph{3D Generation Techniques} Recent advancements in neural generative models have significantly improved the quality and scalability of 3D object generation from single-view reconstruction tasks~\citep{chen2020bsp, mescheder2019occupancy, liu2019learning, wang2018pixel2mesh} to image-conditioned generative models~\citep{melas2023pc2,wu2023sketch,  alliegro2023polydiff, liu2023meshdiffusion, cheng2023sdfusion, zheng2023locally, gupta20233dgen,  muller2023diffrf, zhang20233dshape2vecset}. More recently, methods like Zero123++~\citep{shi2023zero123++}, ImageDream~\citep{wang2023imagedream}, MVDream~\citep{shi2023mvdream}, SyncDreamer~\citep{liu2023syncdreamer} and CAT3D~\citep{gao2024cat3d} leverage multi-view diffusion models to synthesize novel views with high cross-view consistency. By leveraging cross-view attention mechanisms, these approaches enhance 3D reconstruction from sparse input images, resolving depth ambiguities and ensuring globally coherent asset generation. Video-based diffusion models, like ViVid-1-to-3~\citep{kwak2024vivid} and AnimateDiff~\citep{guo2023animatediff}, have also demonstrated the capability to simulate smooth camera trajectories and produce 3D-consistent representations, although with limited flexibility for static 3D object generation.

Similarly, Large Reconstruction Models~\citep{honglrm,wang2023pf} map image tokens to implicit 3D representations such as triplanes or meshes through a transformer using multi-view supervision. By directly optimizing on mesh representations, approaches like InstantMesh~\citep{xu2024instantmesh} and Instant3D~\citep{li2023instant} integrate differentiable isosurface extraction modules to enhance geometric accuracy and reduce memory overhead, achieving efficient and high-quality 3D mesh generation from single images. These frameworks build on scalable transformer architectures, enabling generalization across diverse datasets. Extensions like GRM~\citep{xu2024grm} and MVD2~\citep{zheng2024mvd} explore Gaussian-based representations and mesh-specific optimizations to improve rendering efficiency and surface quality. Despite their advancements, they often struggle with capturing fine details in complex scenes and require carefully curated 3D datasets for robust performance and geometry supervision, unlike the prior multi-view diffusion based approaches. This constrains them to the data distribution they are trained on, unlike the work proposed in this manuscript, where assets can be generated with total freedom of prompts.

Finally, procedural generation techniques, such as those employed in Infinite Photorealistic Worlds~\citep{raistrick2023infinite} and Infinigen Indoors~\citep{raistrick2024infinigen}, utilize algorithmic rules to create diverse and scalable datasets. These systems employ probabilistic programs and node-based tools like Blender to generate photorealistic indoor scenes. 
While highly effective for creating diverse training datasets, procedural approaches are often constrained by predefined content types and can be computationally expensive to enforce certain constraints, such as semantically rich asset integration in indoor environments.

\paragraph{NeRF-based Representations} Neural Radiance Fields (NeRFs) have emerged as a powerful framework for 3D scene representation due to their ability to synthesize novel views with high fidelity~\citep{mildenhall2021nerf}. Extensions such as Zip-NeRF~\citep{barron2023zip} introduce regularization techniques to improve density, sparsity, and texture consistency, facilitating more efficient rendering and reducing artifacts. Advanced methods like Text2NeRF~\citep{zhang2024text2nerf} and Set-the-Scene~\citep{cohen2023set} integrate NeRFs with pre-trained text-to-image diffusion models, enabling text-driven 3D object and scene generation. Despite these advancements, NeRFs are inherently volumetric, making direct integration into standard graphics pipelines challenging. Approaches like NerfMeshing~\citep{rakotosaona2024nerfmeshing} address this limitation by converting radiance fields into explicit 3D meshes introducing a Signed Surface Approximation Network (SSAN) and using techniques such as isosurface extraction.

\paragraph{3D Mesh Generation}
Focusing on mesh-based representations, TextMesh~\citep{tsalicoglou2024textmesh} extends NeRFs to Signed Distance Functions (SDFs) for precise mesh extraction, addressing issues such as oversaturation and texture inconsistency present in earlier methods like DreamFusion~\citep{poole2022dreamfusion}. This comes at the expense of a computational overhead by leveraging multi-view supervision and SDF backbones, where each asset takes around 2 hours to generate. InstantMesh~\citep{xu2024instantmesh}, on the other hand, combines multi-view diffusion models with sparse-view reconstruction frameworks to generate 3D meshes directly from single images. This approach integrates geometric supervision on the mesh surface using differentiable isosurface extraction modules. Despite these advancements, existing mesh generation approaches still struggle with producing highly detailed meshes in complex scenes and maintaining consistent performance across diverse object categories.

\paragraph{Text-to-3D and Synthetic Data for Scene Understanding}
Recent advancements in text-driven 3D generation have evolved from early CLIP-guided methods, such as CLIPMesh~\citep{mohammad2022clip}, to more sophisticated diffusion-based systems leveraging large-scale text-to-image models~\citep{zhang2024text2nerf}. Prompt engineering is critical to improving object placement, spatial alignment, and scene composition. Techniques like ShowRoom3D~\citep{mao2023showroom3d} and SceneScape~\citep{fridman2024scenescape} refine 3D scene synthesis using score distillation and camera trajectory optimization, but often require extensive fine-tuning and are computationally intensive. 

Synthetic data generation helps address the challenge of acquiring annotated datasets for scene understanding~\citep{paulin2023review}, with methods like Text2Room~\citep{hollein2023text2room}, SceneScape~\citep{fridman2024scenescape}, and Ctrl-Room~\citep{fang2023ctrl} using text-to-image models to generate 3D meshes, though they face challenges with geometric distortion, occlusion handling, and scalability. Conditional diffusion models, such as DiffuScene~\citep{tang2024diffuscene}, enhance semantic control over scene composition. Despite these improvements, many methods still struggle with achieving global consistency, geometry fidelity, and artifact-free outputs, especially in complex or occluded scenes. A limitation in these approaches is the dependence on additional 3D supervision, such as geometry through depth or normal maps. In contrast, our method leverages a multi-view Latent Diffusion Model (LDM) pre-trained on RGB instances and camera poses, eliminating the need for extra 3D supervision while maintaining high-quality mesh generation. Additionally, we are not restricted by the domain of the data on which the mesh generation was trained, unlike existing approaches. This grants us the capability to generate assets from previously unseen distributions and enhances the scalability of the later-assembled synthetic rooms. By focusing on generating isolated assets and placing them in existing 3D rooms, we scale existing room data and improve the quality and scalability of synthetic 3D indoor scene generation, bridging the gap between synthetic and real-world data.
\newpage
\section{Prompt Engineering Results}
\label{app:prompts}
\begin{figure}[h!]
\includegraphics[width=\textwidth]{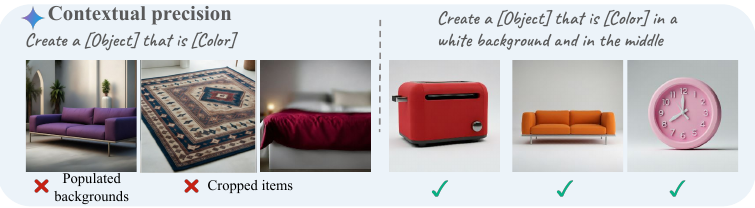}
\caption{Samples of the effect of the contextual precision in the generated outputs.}
\label{fig:results_cont_preci}
\end{figure}

\paragraph{Contextual Precision} Figure~\ref{fig:results_cont_preci} shows examples of the effect of the contextual precision in the generated outputs, as discussed in Section~\ref{sec:methods1}. By adding in ``in an empty white background'' and ``in the middle'', we improve spatial clarity (Right), preventing issues like object cropping or awkward placements (Left).

\paragraph{Scalable Prompt Generations}

A comprehensive explanation of the procedure to generate prompts in a scalable manner is included via a pseudocode in Algorithm~\ref{alg:dataset_creation}, as explained in Section~\ref{sec:methods1}. Table~\ref{tab:prompt_evaluation} shows examples of the different ranges of scores of the resulting prompts together with the corresponding generated asset. 

\begin{algorithm}[H]
\caption{Structured Dataset Creation and Evaluation for 3D Object Text Prompts}
\label{alg:dataset_creation}
\begin{algorithmic}[1]
\REQUIRE Initial category lists: \{\textit{Objects, Materials, Colors, High-level Themes}\}.
\ENSURE Final dataset of ranked and filtered text prompts ready for 3D asset generation.

\STATE \textbf{Step 1: Generate Initial Category Lists}
\STATE \hspace{1em} Prompt LLM to create 10–20 diverse and realistic items for each category.
\STATE \hspace{1em} Organize results into a structured table.

\STATE \textbf{Step 2: Refine and Validate Category Lists}
\STATE \hspace{1em} Remove duplicates and nonsensical combinations from raw lists.

\STATE \textbf{Step 3: Generate Structured Prompt Templates}
\STATE \hspace{1em} Define template variables: \{Object, Color, Material, High-level Theme\}.
\STATE \hspace{1em} Prompt LLM to generate 20 templates with diverse sentence structures.

\STATE \textbf{Step 4: Combine Categories to Create Text Prompts}
\STATE \hspace{1em} Randomly sample combinations of \{Object, Material, Color, High-level Theme\}.
\STATE \hspace{1em} Fill placeholders in structured templates with sampled values.

\STATE \textbf{Step 5: Evaluate and Refine Prompts}
\STATE \hspace{1em} Feed generated prompts into LLM for evaluation based on:
\STATE \hspace{2em} (a) \textbf{Coherence:} Does the object-material-color combination make sense?
\STATE \hspace{2em} (b) \textbf{Specificity:} Is the object and setting clearly described?
\STATE \hspace{2em} (c) \textbf{Creativity:} Does the prompt inspire visually compelling designs?
\STATE \hspace{1em} Assign a realism score (1–10) to each prompt, with explanations.

\STATE \textbf{Step 6: Rank and Output Results}
\STATE \hspace{1em} Sort prompts by realism scores and output resulting list as \texttt{csv} file.

\RETURN Final dataset of filtered and ranked prompts.
\end{algorithmic}
\end{algorithm}

\begin{table}[h!]
\centering
\caption{Samples of prompts, scores, and corresponding figures resulting from the proposed prompt engineering approach. The explanations behind the scores given by the \textit{Gemini} are: \textit{Coherence} - Materials and colors are plausible for the object, \textit{Specificity} - Clarity of the design details, and \textit{Creativity} - Suggests visually compelling design elements.}

\begin{tabular}{p{5.1cm}p{2.9cm}m{5cm}}
\toprule
\textbf{Prompt} & \textbf{Score} & \textbf{Mesh} \\ \midrule

\textit{Create a green toaster, suitable for a Japanese environment  in an empty white background and in the middle.} & \textbf{9.0}: \hspace{1.75cm} Coherence 9/10, Specificity 9/10, Creativity \hspace{0.65cm}  9/10. & \includegraphics[width=\hsize]{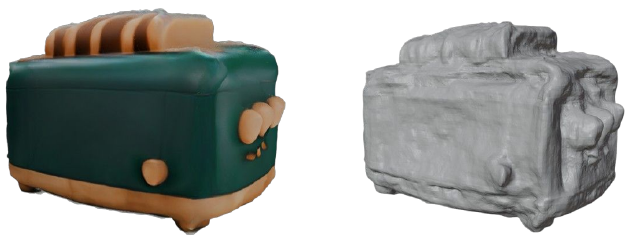} \\\midrule
\textit{An antique concept: a white armchair that seamlessly integrates into the design in an empty white background and in the middle.} & \textbf{8.0}: \hspace{1.75cm}Coherence 9/10, Specificity 7/10, Creativity \hspace{0.65cm} 8/10. & \includegraphics[width=\hsize]{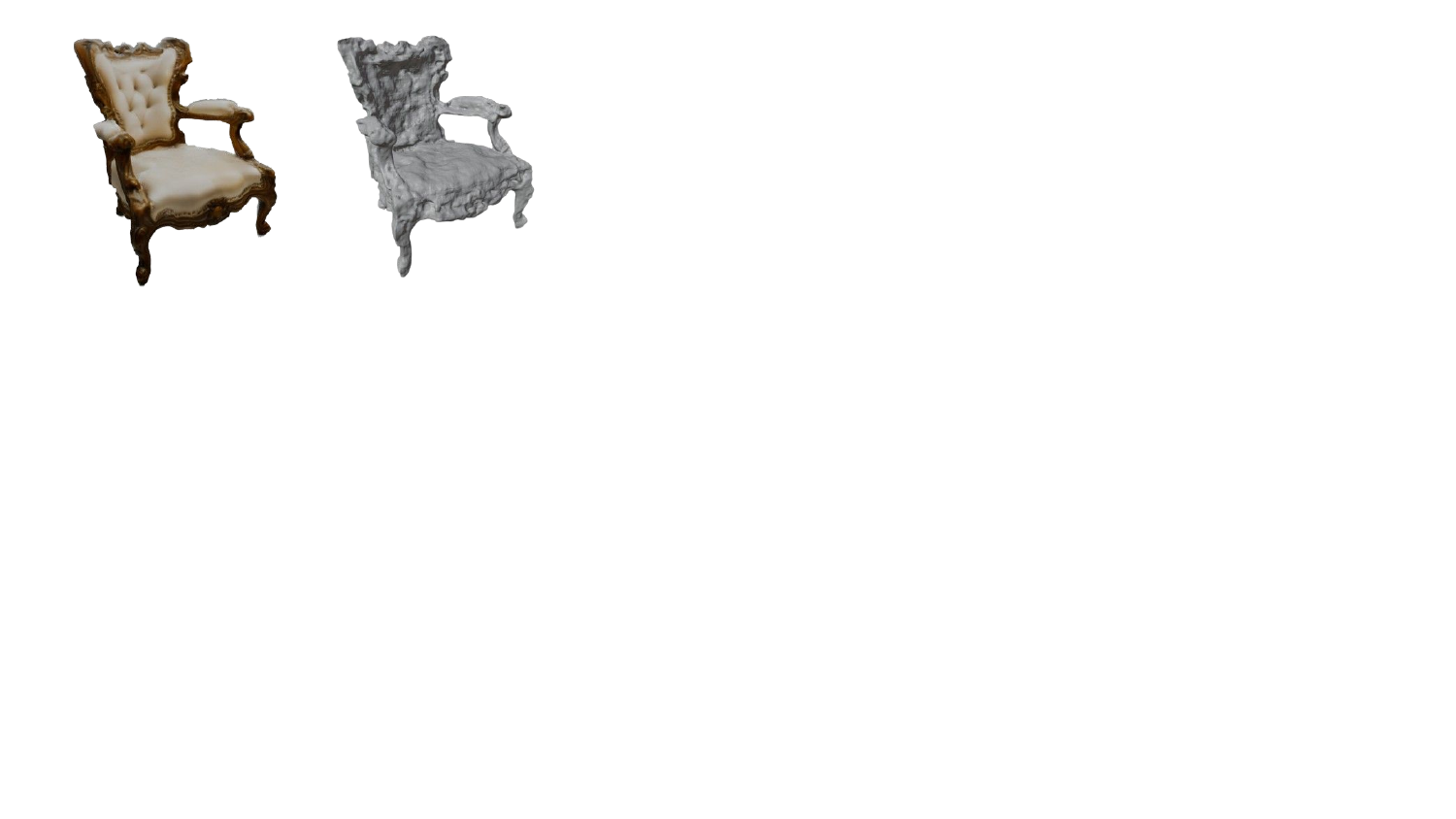} \\ \midrule
\textit{An industrial interpretation of a plant, realized in pink ceramic in an empty white background and in the middle.} & \textbf{7.0}: \hspace{1.75cm}Coherence 5/10, Specificity 9/10, Creativity \hspace{0.65cm} 7/10. & \includegraphics[width=\hsize]{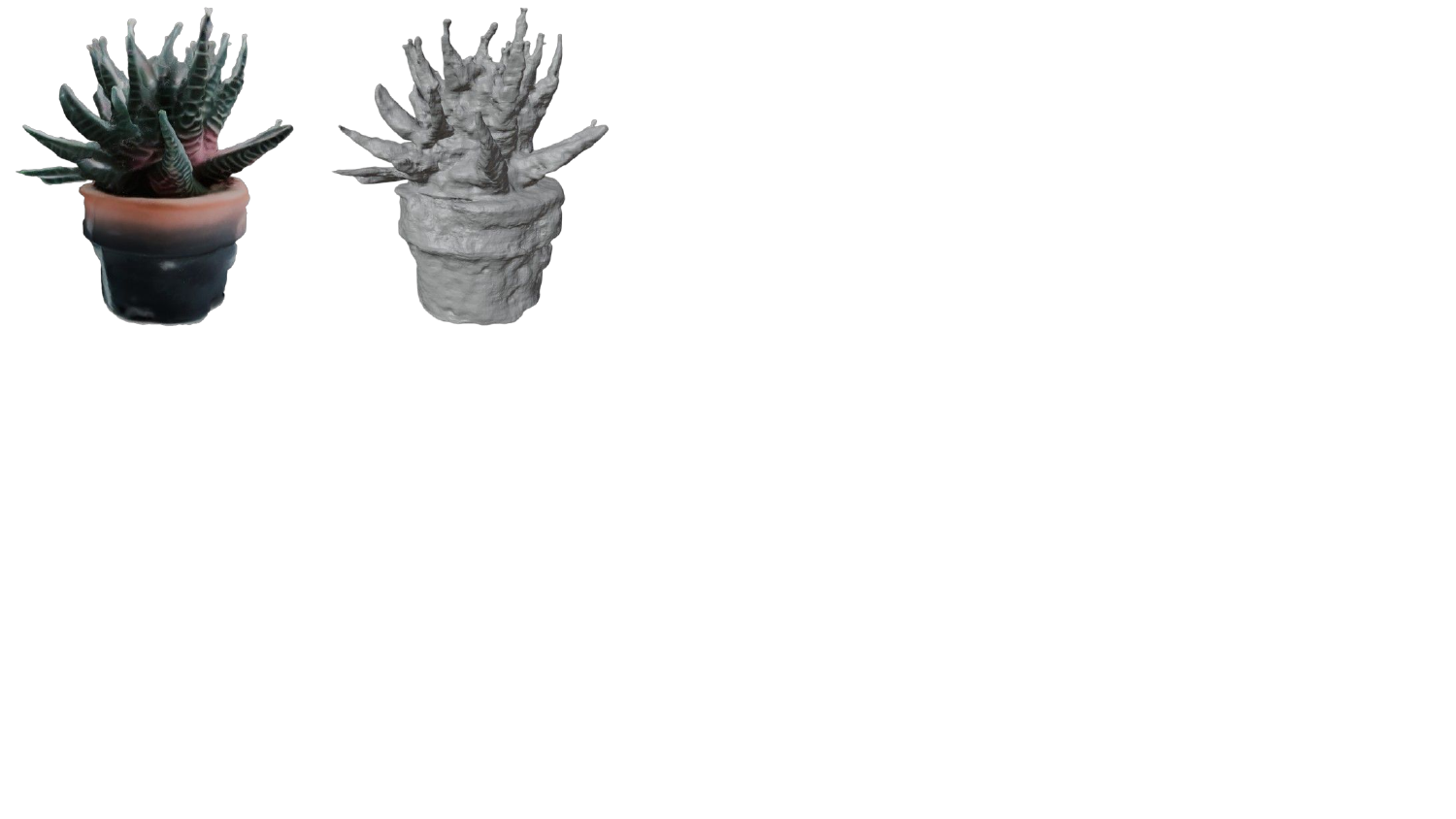} \\ \midrule
\textit{Design a vase that is black and crafted from cotton reflecting the core principles of Art Deco design in an empty white background and in the middle. }& \textbf{5.4}: \hspace{1.75cm}Coherence 3/10, Specificity 9/10, Creativity \hspace{0.65cm} 4/10. & \hspace{0.45cm} \includegraphics[width=0.8\hsize]{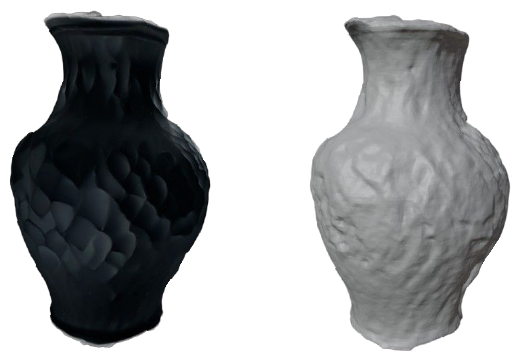} \\ \bottomrule

\end{tabular}

\label{tab:prompt_evaluation}
\end{table}

\vspace*{7cm}
\section{Additional Qualitative Results}
\label{app:results}
\begin{figure}[h]
\vspace*{-0.4cm}
\centering
\includegraphics[width=0.74\linewidth]{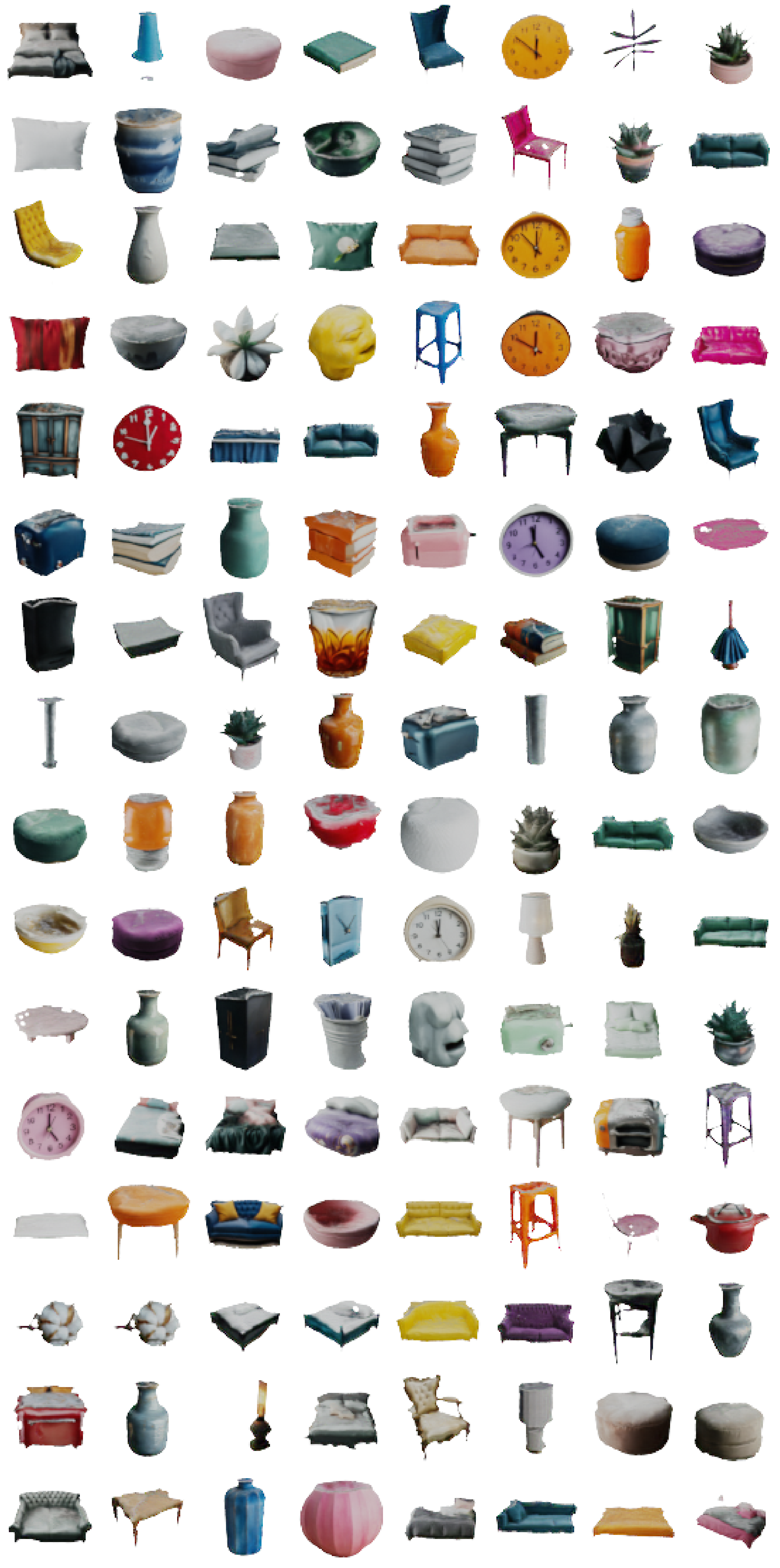}
\vspace*{-0.4cm}
\caption{Visualization of generated 3D objects. Note samples of failure cases where holes are present due to reconstruction errors.}
\label{fig:examples_assets}
\end{figure}

\end{document}

%% file: math_commands.tex
\usepackage{amsmath,amsfonts,bm}

\def\eqref#1{equation~\ref{#1}}

\def\1{\bm{1}}

\DeclareMathAlphabet{\mathsfit}{\encodingdefault}{\sfdefault}{m}{sl}
\SetMathAlphabet{\mathsfit}{bold}{\encodingdefault}{\sfdefault}{bx}{n}

